# Development of a Multi-User Recognition Engine for Handwritten Bangla Basic Characters and Digits

Sandip Rakshit [1], Debkumar Ghosal[2], Tanmoy Das[2], Subhrajit Dutta[2], Subhadip Basu[2]

[1] Techno India College of Technology, Kolkata, India
[2] Computer Science and Engineering Department, Jadavpur University, India

[1] Corresponding author. E-mail: *rakshitsandip@ieee.org*

**Abstract:** The objective of the paper is to recognize handwritten samples of basic Bangla characters using Tesseract open source Optical Character Recognition (OCR) engine under Apache License 2.0. Handwritten data samples containing isolated Bangla basic characters and digits were collected from different users. Tesseract is trained with user-specific data samples of document pages to generate separate user-models representing a unique language-set. Each such language-set recognizes isolated basic Bangla handwritten test samples collected from the designated users. On a three user model, the system is trained with 919, 928 and 648 isolated handwritten character and digit samples and the performance is tested on 1527, 14116 and 1279 character and digit samples, collected form the test datasets of the three users respectively. The user specific character/digit recognition accuracies were obtained as 90.66%, 91.66% and 96.87% respectively. The overall basic character-level and digit level accuracy of the system is observed as 92.15% and 97.37%. The system fails to segment 12.33% characters and 15.96% digits and also erroneously classifies 7.85% characters and 2.63% on the overall dataset.

## 1. Introduction

Optical Character Recognition (OCR) is still an active area of research, especially for handwritten text. Success of the commercially available OCR system is yet to be extended to handwritten text. It is mainly due to the fact that numerous variations in writing styles of individuals make recognition of handwritten characters difficult. Past work on OCR of handwritten alphabet and numerals has been mostly found to concentrate on Roman script [1-2], related to English and some European languages, and scripts related to Asian languages like Chinese, Korean, and Japanese.

Among Indian scripts, Devnagri, Tamil, Oriya and Bangla have started to receive attention for OCR related research in the recent years. Out of these, Bangla is the second most popular script and language in the Indian subcontinent. As a script, it is used for Bangla, Ahamia and Manipuri languages. Bangla, which is also the national language of Bangladesh, is the fifth most popular language in the world. So is the importance of Bangla both as a script and as a language. But evidences of research on OCR of handwritten Bangla characters, as observed in the literature, are a few in numbers.

Not only because of numerous variation of writing styles of different individuals but also for the complex nature of Bangla alphabet, automatic recognition of handwritten Bangla characters still poses some potential problems to the researchers. Compared to Roman alphabet, basic Bangla alphabet consists of a much larger number of characters. The number of characters in basic Bangla alphabet is 50. And some characters therein resemble pair wise so closely that the only sign of small difference left between them is a period or a small line. Handwritten samples of all 50 symbols of basic Bangla alphabet are shown in Fig 1.

Development of a handwritten OCR engine with high recognition accuracy is a still an open problem for the research community. Lot of research efforts have already been reported [1-8] on different key aspects of handwritten character recognition systems. In the current work, instead of developing a new handwritten OCR engine from scratch, we have used Tesseract 2.01 [9], an open source OCR Engine under Apache License 2.0, for recognition of handwritten pages consisting of basic bangle characters. Tesseract OCR engine provides high level of character recognition accuracy on poorly printed or poorly copied dense text. In our earlier works [10-11], we had developed a system for estimation of recognition accuracy of Tesseract OCR engine on handwritten lower case Roman character samples, collected from a single/multiple user. But the performance of this OCR engine could not be tested extensively on handwriting samples of complex Indic scripts. This has been one of the major motivations behind the current work, presented in this paper.

In the current work, we have used Tesseract to perform user specific training on handwriting samples of basic Bangla characters. The performance is evaluated on the categories of document pages for observation of character level accuracies only.

(a) Handwritten samples of Bangla vowels

(b) Handwritten samples of Bangla consonants

(c) Handwritten samples of Bangla digits

**Fig. 1(a-c).** Handwritten Basic character and digit samples of Bangla script.

The developed system is having potential utilities in designing fast user specific online handwriting recognition systems using pen based input devices.

## 2. Overview Of The Tesseract Ocr Engine

Tesseract is an open source (under Apache License 2.0) offline optical character recognition engine, originally developed at Hewlett Packard from 1984 to 1994. Tesseract was first started as a PhD research project in HPLabs, Bristol [12]. In the year 1995 it is sent to UNLV where it proved its worth against the commercial engines of the time [13]. In the year 2005 Hewlett Packard and University of Nevada, Las Vegas, released it. Now it is partially funded by Google [14] and released under the Apache license, version 2.0. The latest version, Tesseract 2.03 is released in April, 2008. In the current work, we have used Tesseract version 2.01, released in August 2007.

Like any standard OCR engine, Tesseract is developed on top of the key functional modules like, line and word finder, word recognizer, static character classifier, linguistic analyzer and an adaptive classifier. However, it does not support document layout analysis, output formatting and graphical user interface. Currently, Tesseract can recognize printed text written in English, Spanish, French, Italian, Dutch, German and various other languages.

For exampleTo train Tesseract in Bangla language 8 data files are required in tessdata sub directory. The 8 files used for Bangla are to be generated as follows:

    tessdata/ban.freq-dawg
    tessdata/ban.word-dawg
    tessdata/ban.user-words
    tessdata/ban.inttemp
    tessdata/ban.normproto
    tessdata/ban.pffmtable
    tessdata/ban.unicharset
    tessdata/ban.DangAmbigs

## 3. The Present Work

In the current work, Tesseract 2.01 is used for recognition of handwriting samples of isolated basic characters and digits of Bangla script. Key functional modules of the developed system are discussed the following sub-sections.

### 3.1. Collection of the dataset

For collection of the dataset in the current experiment, we have concentrated on basic character and digits of Bangla script. We have used a pen based input device (stylus/tablet), manufactured by i-Ball (as shown in Fig. 2a) and collected handwritten character samples from different users. Three different datasets are collected from each of the three users of the designed system. In the first dataset (Dataset-1) isolated hand written Bangla Vowel characters collected, as shown in figure 1(a). In the second dataset (Dataset-2) hand written Bangla Consonant Characters as shown in figure 1(b) and in the third dataset (Dataset -3) hand written Bangla digits were collected from each user. We also consider here 3 user   for collecting the data and testing also for training phase first we also
      write here user specific data (both Bangla characters & digits).

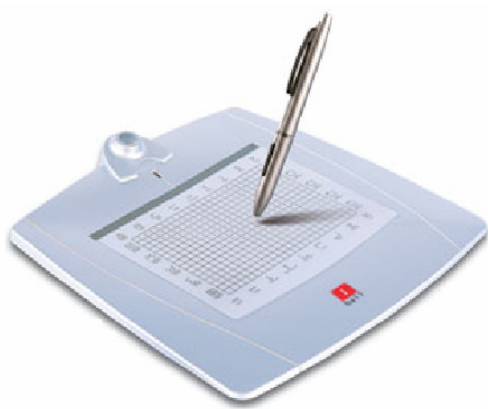

(a)

(b)          (c)

**Fig. 2(a).** The pen based input device. **(b-c)** Sample document pages containing training sets of isolated characters and free flow text

**Table 1.** Composition of the training and test set character samples for different users

|  |  | Data set-1 | Data set-2 | Data set-3 | Overall |
|---|---|---|---|---|---|
| User 1 | Train set | 154 | 595 | 170 | 919 |
|  | Test set | 308 | 939 | 280 | 1527 |
| User 2 | Train set | 298 | 490 | 140 | 928 |
|  | Test set | 333 | 770 | 308 | 1411 |
| User 3 | Train set | 198 | 350 | 100 | 648 |
|  | Test set | 363 | 596 | 320 | 1279 |

The overall distribution of the character samples in the training and the test sets for the three users is shown in Table 1.

### 3.2. Labeling training data

For labeling the training samples using Tesseract we have taken help of a tool named bbTesseract [13]. To generate the training files for a specific user, we need to prepare the box files for each training images using the following command:

*tesseract fontfile.tif fontfile batch.nochop makebox*

The box file is a text file that includes the characters in the training image, in order, one per line, with the coordinates of the bounding box around the image. The new Tesseract 2.01 has a mode in which it will output a text file of the required format. Some times the character set is different to its current training, it will naturally have the text incorrect. In that case we have to manually edit the file (using bbTesseract) to correct the incorrect characters in it. Then we have to rename fontfile.txt to fontfile.box. Fig. 3 shows a screenshot of the bbTesseract tool, used for labeling the training set.

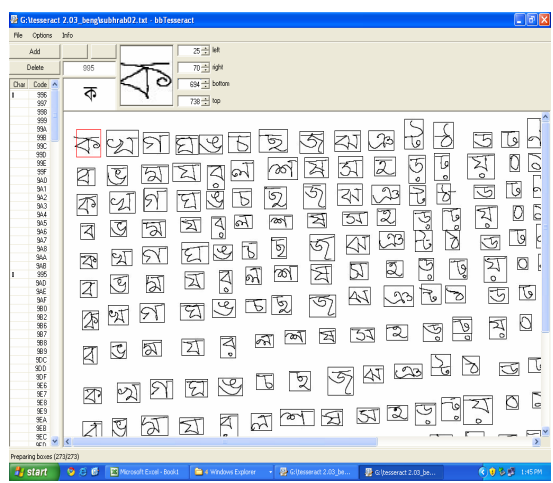

**Fig.3.** A sample screenshot of the bbTesseract tool

### 3.3. Training the data using Tesseract OCR engine

For training a new handwritten character set for any user, we have to put in the effort to get one good box file for a handwritten document page, run the rest of the training process, discussed below, to create a new language set. Then use Tesseract again using the newly created language set to label the rest of the box files corresponding to the remaining training images using the process discussed in section 3.2.

For each of our training image, boxfile pairs, run Tesseract in training mode using the following command:

*tesseract fontfile.tif junk nobatch box.train*

The output of this step is fontfile.tr which contains the features of each character of the training page. The character shape features can be clustered using the mftraining and cntraining programs:

*mftraining fontfile_1.tr fontfile_2.tr ...*

This will output three data files: inttemp , pffmtable and Microfeat, and the following command:

*cntraining fontfile_1.tr fontfile_2.tr ...*

This will output the normproto data file. Now, to generate the unicharset data file, unicharset_extractor program is used as follows:

*unicharset_extractor fontfile_1.box fontfile_2.box ...*

Tesseract uses 3 dictionary files for each language. Two of the files are coded as a Directed Acyclic Word Graph (DAWG), and the other is a plain UTF-8 text file. To make the DAWG dictionary files a wordlist is required for our language. The wordlist is formatted as a UTF-8 text file with one word per line. The corresponding command is:

*wordlist2dawg frequent_words_list freq-dawg*
*wordlist2dawg words_list word-dawg*

The third dictionary file name is user-words and is usually empty. The final data file of Tesseract is DangAmbigs file. This file cannot be used to translate characters from one set to another. The DangAmbigs file may be empty also.

Now we have to collect all the 8 files and rename them with a lang. prefix, where lang is the 3-letter code for our language and put them in our tessdata directory. Tesseract can then recognize text in our language using the command:

*tesseract image.tif output -l lang*

## 4. Experimental Results

For conducting the current experiment, three user-specific language sets are generated using Tesseract open source OCR engine. The training and test patterns of each individual user are spread over two types of datasets, as described in Sec. 3.1. The experiment is focused on testing the segmentation and core recognition accuracy of Tesseract OCR engine on free flow handwritten annotations written using digital pens by different users. The linguistic analysis module of Tesseract, involving the language files freq-dawg, word-dawg, user-words and DangAmbigs are not utilized in the current experiment. To evaluate the performance of the present technique the following expression is developed.

Recognition accuracy = $(CB_{tB} / (CB_{mB} + CB_{sB})) * 100$

where $CB_{tB}$ = the number of character segments producing true classification result and $CB_{mB}$ = the number of misclassified character segments and $CB_{sB}$ signifies the number of character Tesseract fails to segment, i.e., producing under segmentation. The rejected character/word samples are excluded from computation of recognition accuracy of the designed system.

Table 2(a-c) shows an analysis of successful classification (SC), misclassification (Misc), segmentation failure (SF) and rejection (Rej) results on the test samples of the three users. Fig. 4(a-c) shows character wise distribution of success and failure accuracies on the three test datasets for all users combined. As observed from the experimentation a significant proportion rejection cases evolve out of the word segmentation failures. This is so because Tesseract is originally designed to recognize printed document pages with uniformity in baseline and character/word spacings. Another source of error is due to the internal segmentation of some of the characters.

On a three user model, the system is trained with 919, 928 and 648 isolated handwritten character and digit samples and the performance is tested on 1527, 14116 and 1279 character and digit samples, collected form the test datasets of the three users respectively. The user specific character/digit recognition accuracies were obtained as 90.66%, 91.66% and 96.87% respectively. The overall basic character-level and digit level accuracy of the system is observed as 92.15% and 97.37%. The system fails to segment 12.33% characters and 15.96% digits and also erroneously classifies 7.85% characters and 2.63% on the overall dataset.

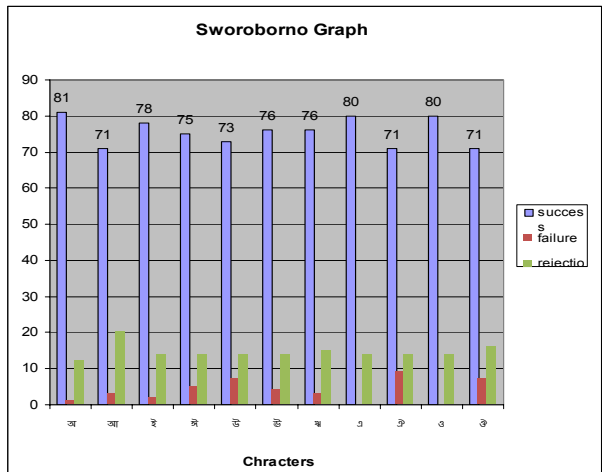

**Fig. 4(a).** Distribution of success and failure cases over the test samples of Dataset-1

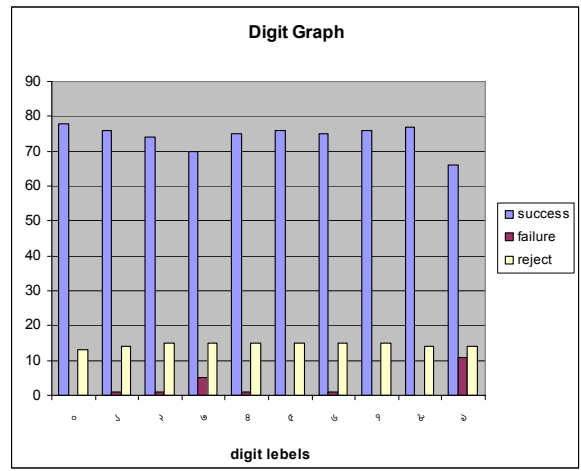

**Fig. 4(b).** Distribution of success and failure cases over the test samples of dataset-3

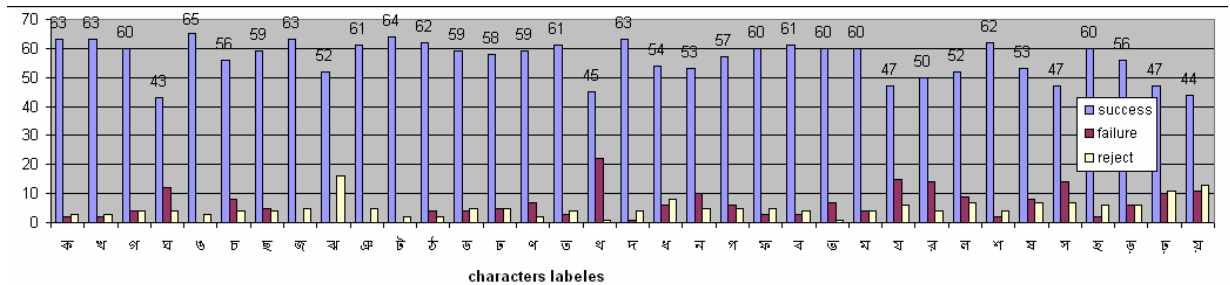

**Fig. 4(c).** Distribution of success and failure cases over the test samples of Dataset-2

**Table 2.** Analysis of recognition performance of the developed system

(a) Recognition performance of User-1 test dataset

|      | Dataset-1 | Dataset-2 | Dataset-3 | Overall |
|------|-----------|-----------|-----------|---------|
| **SC**   | 92.27 | 88.34 | 93.93 | 90.66 |
| **Misc** | 7.73  | 11.66 | 6.07  | 9.34  |
| **Rej**  | 11.68 | 6.92  | 23.57 | 10.93 |

(b) Recognition performance of User-2 test dataset

|      | Dataset-1 | Dataset-2 | Dataset-3 | Overall |
|------|-----------|-----------|-----------|---------|
| **SC**   | 94.36 | 88.52 | 97.68 | 91.66 |
| **Misc** | 5.64  | 11.48 | 2.32  | 8.34  |
| **Rej**  | 23.72 | 3.76  | 15.90 | 11.12 |

(c) Recognition performance of User-3 test dataset

|      | Dataset-1 | Dataset-2 | Dataset-3 | Overall |
|------|-----------|-----------|-----------|---------|
| **SC**   | 98.73     | 94.45     | 99.65     | 96.87   |
| **Misc** | 1.27      | 5.55      | 0.35      | 3.13    |
| **Rej**  | 12.67     | 3.35      | 9.38      | 7.50    |

## 5. Conclusion

As observed from the experimental results, Tesseract OCR engine fares reasonably with respect to the core recognition accuracy on user-specific handwritten samples of basic Bangla characters and digits. The performance of the system is validated on a three-user recognition engine. A major drawback of the current technique is its failure to avoid over-segmentation in some of the characters. Also the system fails to segment cursive words in many cases leading to under-segmentation and rejection. The performance of the designed system may be improved by incorporating more training samples for each user and inclusion of word-level dictionary matching techniques.

## Acknowledgements

One of the authors, Mr. Sandip Rakshit is thankful to the authorities of Techno India College of Technology for necessary supports during the research work. Dr. Subhadip Basu is thankful to the "Center for Microprocessor Application for Training Education and Research", "Project on Storage Retrieval and Understanding of Video for Multimedia" of Computer Science & Engineering Department, Jadavpur University, for providing infrastructure facilities during progress of the work.